\documentclass[runningheads]{llncs}

\usepackage{eccv}
\usepackage{graphicx}
\usepackage{booktabs}
\usepackage[accsupp]{axessibility}
\usepackage{hyperref}
\usepackage{orcidlink}
\usepackage{multirow}
\usepackage{pifont}
\usepackage{marvosym}
\usepackage{placeins}
\usepackage{float}
\newcommand{\myheading}[1]{\par\noindent\textbf{#1}\space}

\begin{document}

\title{EditVerse3D: High-Quality 3D Object Editing with Region-Aware Learning}
\titlerunning{EditVerse3D: High-Quality 3D Object Editing}

\author{Youtan Yin\inst{1,2}\textsuperscript{(\Letter)}\orcidlink{0000-0002-7321-7927} \and
Yanning Zhou\inst{2} \and
Jiacheng Wei\inst{1} \and
Xiaofeng Yang\inst{1} \and
Jun Zhang\inst{2} \and
Jiayang Bai\inst{2} \and
Jingwen Ye\inst{2} \and
Weidong Zhang\inst{2} \and
Guosheng Lin\inst{1}
}

\authorrunning{Y.~Yin et al.}
\institute{College of Computing and Data Science, Nanyang Technological University,\\
Singapore, Singapore \email{\{youtan001,xiaofeng001\}@e.ntu.edu.sg, weijiacheng.gaw@gmail.com, gslin@ntu.edu.sg}\\
\and
Tencent AIPD, Shenzhen, China\\
\email{ynzhou@cse.cuhk.edu.hk, \{junejzhang,darcybai,jingwenye,wadewdzhang\}@tencent.com}
}

\maketitle

\begin{abstract}
Local editing of 3D objects remains a long-standing challenge.
When interacting with 3D content, humans naturally tend to specify a coarse region of interest for modification rather than defining precise editing boundaries. However, previous methods rely on fully edited 2D images, precise 3D masks, or redundant pipelines, which present a gap.
To bridge this gap, we propose EditVerse3D, a novel 3D editing framework that enables high-quality object editing under such coarse guidance. Our approach takes as input a 3D object to be edited, a coarse 3D bounding box indicating the target region, and a reference 2D image describing the desired modification.
It produces a coherent, high-fidelity edited 3D object. To facilitate this editing, we introduce a novel region-aware adaptive loss that emphasizes hard-to-learn regions and balances the objective between target and preserved areas.
Complementing our loss function, we enhance model robustness and generalization through targeted data augmentations, such as training with scaled 3D masks and filtering out unrealistic editing pairs.
We construct a large-scale 3D editing dataset derived from parts information.
Extensive experiments demonstrate that EditVerse3D achieves superior visual quality and quantitative performance compared to existing 3D editing approaches. Please visit our project page at \url{https://editverse3d.github.io}.
\end{abstract}

\section{Introduction}

Local editing of 3D objects has been challenging due to the inherent complexity of the 3D context compared to 2D. However, the broad applicability of 3D editing has attracted significant attention from the research community \cite{yan3DSceneEditorControllable3D2026,zhangAdvancing3DGaussian2025,hanARAPGSDragdrivenAsRigidAsPossible2026,sellaBlendedPointCloud2025,fangChatEdit3DInteractive3D2025,huCraftMeshHighFidelityGenerative2026,heCTRLDControllableDynamic2025,rojasDATENeRFDepthAwareTextBased2025,xiaScalableConsistent3D2026}. Existing approaches can generally be divided into three categories.

\begin{figure}[htb!]
    \centering
    \includegraphics[width=\textwidth]{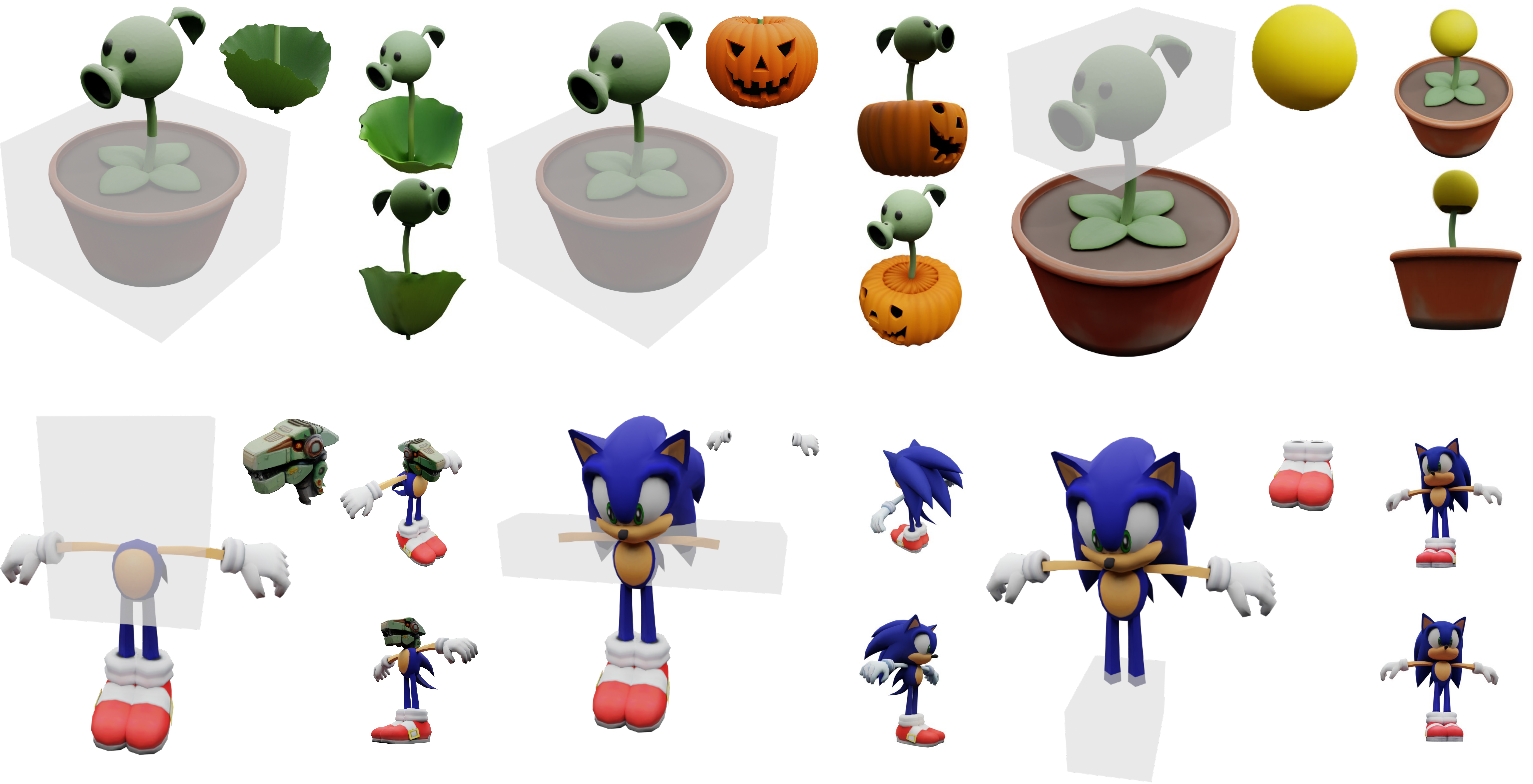}
    \caption{\textbf{Editing results of our method.} Given a 3D object, a user-specified coarse 3D bounding box indicating the target editing region, and an image prompt defining the editing goal, our approach generates high-quality, coherent edits. Our method does not require fully edited 2D views, precise 3D masks, or redundant pipelines.}
    \label{fig:teaser}
\end{figure}

The first approach \cite{liCMDControllableMultiview2025,chenGeneric3DDiffusion2024,qiTailor3DCustomized3D2024,bardaInstant3ditMultiviewInpainting2025,gao3DMeshEditing2025,kimDreamCatalystFastHighQuality2025,xuGGEditorLocallyEditing2024,erkocPrEditor3DFastPrecise2025,yinORNeRFObjectRemoving2023} involves rendering multiple 2D views of the input 3D object, editing these views, and subsequently lifting the edits back into 3D. However, this method faces two significant challenges. First, the back-and-forth between 3D and 2D introduces cumulative errors, resulting in a decline in editing quality. Second, editing in the 2D context ignores 3D structure, often leading to inconsistencies across multiple views. Moreover, although large vision models (LVMs) \cite{rombachHighResolutionImageSynthesis2022,lipmanFlowMatchingGenerative2023,blackforestlabsFLUX1KontextFlow2025,wuQwenImageTechnicalReport2025,googleGoogleGemini2025} have achieved remarkable results, precise 2D edits \cite{juBrushNetPlugandPlayImage2024,renFDSFrequencyAwareDenoising2025,dengFireFlowFastInversion2025,zhangEnablingInstructionalImage2025,zhouMultiturnConsistentImage2025,mengSDEditGuidedImage2022,routSemanticImageInversion2025,hongGeneralizedTrainingFreeTextGuided2025} remain a considerable challenge.
Representative editing results from our method are shown in Fig.~\ref{fig:teaser}.

\FloatBarrier

The second approach \cite{sellaVoxETextGuidedVoxel2023,chenSHAPEDITORInstructionGuidedLatent2024,dinhGeometryStyle3D2025,dongInteractive3DCreateWhat2024,liuMakeYour3DFastConsistent2024,zhuangTIPEditorAccurate3D2024,chenPlasticine3D3DNonRigid2024,lePreservingIdentityVariational2024,mikaeiliSKEDSketchguidedTextbased2023,liuSketchDreamSketchbasedTextTo3D2024} is based on the Score Distillation Sampling (SDS) \cite{pooleDreamFusionTextto3DUsing2023} scheme and its variants, which use 2D models to guide the 3D reconstruction process toward the edited 3D representation. However, these methods are computationally intensive, time-consuming, struggle with network optimization, and often fail to produce high-quality results.

Recently, the development of 3D generative models \cite{yangHunyuan3D10Unified2024,zhaoHunyuan3D20Scaling2026,laiHunyuan3D25HighFidelity2025,teamhunyuan3dHunyuan3DOmniUnifiedFramework2025,xiangStructured3DLatents2025,zhang3DShape2VecSet3DShape2023,yaoCASTComponentAligned3D2025,wuDirect3DS2Gigascale3D2025,liMeshPadInteractiveSketchConditioned2025,liStep1X3DHighFidelityControllable2025} has shown great promise, achieving impressive results in 3D generation tasks. Some methods have explored training-free 3D editing \cite{liVoxHammerTrainingFreePrecise2026,parelli3DLATTELatentSpace2026,yeNano3DTrainingFreeApproach2026} based on 3D generative models, which modify the inference process to output an edited 3D object conditioned on editing instructions. These methods typically adapt training-free 2D editing techniques \cite{lugmayrRePaintInpaintingUsing2022,wangTamingRectifiedFlow2025,kulikovFlowEditInversionFreeTextBased2025} developed for 2D generative models to the 3D models. However, due to the increased complexity of 3D objects, their generalization ability is limited, and they often work only in specific cases.

Leveraging advances in 3D generative models, we propose an end-to-end 3D editing framework that overcomes the limitations of existing methods that rely on complex, multi-stage pipelines. Our approach eliminates the need for additional inputs such as pre-edited 2D views, 2D masks, or precise 3D masks, while maintaining the efficiency of 3D generative models to produce high-quality results within minutes. The framework takes three key inputs: \textbf{(1)} the 3D object to be edited, \textbf{(2)} a coarse 3D bounding box specifying the target region, and \textbf{(3)} a 2D image defining the editing goal. Using these inputs, the model directly generates the edited 3D object in a streamlined manner.

Specifically, we build our 3D editing model based on the 3D generative backbone, TRELLIS \cite{xiangStructured3DLatents2025}. We carefully design the model architecture (\cref{subsec:network architecture}) and training strategy (\cref{subsec:training strategy}) to preserve the original model's generative capabilities while adapting it to support our input format. To train our model for high-quality edits under relatively loose inputs, we introduce a loss strategy that adaptively forces the model to focus more on the poorly learned region. Moreover, data augmentation techniques are applied to improve the model's generalization.

Due to the unavailability of large-scale 3D editing datasets tailored to our setting, we construct a comprehensive dataset of 3D editing pairs using 3D segmentation information (\cref{subsec:dataset curation}). The construction process follows a straightforward approach: we remove a part from a 3D object and treat its restoration as an ``add'' editing operation. Specifically, the 3D object after removal serves as the input, while the original 3D object acts as the ground truth. The removed part is used as the 3D mask, which is rendered to generate the corresponding 2D image prompt. Although this dataset primarily consists of ``add'' edits, our experiments demonstrate that models trained on it can generalize effectively to ``replace'' edits. Conceptually, ``add'' can be regarded as a special case of ``replace,'' where the target region lacks any existing 3D structure.

To address the scarcity of 3D segmentation datasets, we further expand our dataset by incorporating data from existing 3D object repositories, such as Objaverse \cite{deitkeObjaverseUniverseAnnotated2023,deitkeObjaverseXLUniverse10M2023}, which contain objects naturally composed of multiple parts. As a result, we curate an extensive 3D editing dataset comprising approximately 85k meshes and 500k editing pairs.

In summary, our method has the following contributions:
\begin{itemize}
    \item An end-to-end 3D editing framework with adaptive loss reweighting and data augmentation, enabling robust generalization and practical usability under coarse input settings.
    \item A large-scale 3D editing dataset that addresses the long-standing challenge of lacking supervised training data for 3D editing.
    \item A novel 3D editing pipeline that produces high-quality results, outperforming previous methods in both visual quality and quantitative assessment.
\end{itemize}

\clearpage
\section{Related Works}

\subsection{3D Generation}

With the rapid advances in diffusion \cite{rombachHighResolutionImageSynthesis2022} and flow-matching models \cite{lipmanFlowMatchingGenerative2023} for language and 2D vision tasks, researchers have increasingly adapted these architectures for 3D generation \cite{choi3DoodleCompactAbstraction2024,zhang3DShape2VecSet3DShape2023,wuAmodal3RAmodal3D2025,yaoCASTComponentAligned3D2025,zhengConstructing3DScene2025,wuDirect3DS2Gigascale3D2025,lanGaussianAnythingInteractivePoint2025,xuGRMLargeGaussian2025,linKiss3DGenRepurposingImage2025,songTopologyPreservedAutoregressiveMesh2026,liMeshPadInteractiveSketchConditioned2025,wangMoGeUnlockingAccurate2025,luMOVISEnhancingMultiObject2025,weiOctGPTOctreebasedMultiscale2025,luOrientationMattersMaking2025,yePrimitiveAnythingHumanCrafted3D2025,bokhovkinSceneFactorFactoredLatent2025,heSparseFlexHighResolutionArbitraryTopology2025,liStep1X3DHighFidelityControllable2025,liTextto3DGeneration2D2025,chenUltra3DEfficientHighFidelity2025,laiUnleashingVecsetDiffusion2025,longWonder3DSingleImage2024}, achieving remarkable results. The development of large-scale 3D generative models has established a strong foundation for downstream applications, accompanied by the emergence of extensive 3D object datasets \cite{deitkeObjaverseUniverseAnnotated2023,deitkeObjaverseXLUniverse10M2023,xiangStructured3DLatents2025,zhangTexVerseUniverse3D2025}. Furthermore, many studies have focused on 3D segmentation-related tasks, such as jointly generating 3D content and its corresponding semantic segmentation \cite{tangEfficientPartlevel3D2025,yangOmniPartPartAware3D2025,linPartCrafterStructured3D2025,chenPartGenPartlevel3D2025,dongOneMoreContextual2025}, or directly segmenting 3D assets \cite{liuPartFieldLearning3D2025,yangSAMPart3DSegmentAny2024,zhuPartSAMScalablePromptable2026,yanXPartHighFidelity2025,zhaoAssemblerScalable3D2025,maFindAnyPart2025,yangHoloPartGenerative3D2026,chenReasoning3DGroundingReasoning2024,fischerSAMaMaterialAware3D2026,cenSegmentAny3D2025,tangSegmentAnyMesh2024,gaoSelfsupervisedLearningHybrid2025}. These tasks typically rely on 3D segmentation datasets \cite{dongOneMoreContextual2025,yangSAMPart3DSegmentAny2024,maP3SAMNative3D2025,linPartCrafterStructured3D2025}, further bridging the gap toward 3D editing. Consequently, with the progress in 3D generative modeling and the availability of large-scale 3D data, supervised training of 3D editing models has become increasingly feasible.

\subsection{3D Editing}

3D editing has long been an actively explored research topic, yet there remains substantial room for improvement. In the absence of large-scale 3D generative models, earlier approaches \cite{liCMDControllableMultiview2025,bar-onEditP233DEditing2025,chenGeneric3DDiffusion2024,qiTailor3DCustomized3D2024,bardaInstant3ditMultiviewInpainting2025,gao3DMeshEditing2025,kimDreamCatalystFastHighQuality2025,xuGGEditorLocallyEditing2024,erkocPrEditor3DFastPrecise2025,yinORNeRFObjectRemoving2023} typically render several 2D views, edit the rendered images, and then lift the results back to 3D. Naturally, this 3D--2D--3D process accumulates errors and often leads to poor 3D consistency. Other methods \cite{sellaVoxETextGuidedVoxel2023,chenSHAPEDITORInstructionGuidedLatent2024,dinhGeometryStyle3D2025,dongInteractive3DCreateWhat2024,liuMakeYour3DFastConsistent2024,zhuangTIPEditorAccurate3D2024,chenPlasticine3D3DNonRigid2024,lePreservingIdentityVariational2024,mikaeiliSKEDSketchguidedTextbased2023,liuSketchDreamSketchbasedTextTo3D2024} adopted score distillation sampling (SDS) \cite{pooleDreamFusionTextto3DUsing2023} to optimize 3D representations under the guidance of 2D priors, but such approaches are usually hard to train and yield limited visual quality. With the emergence of 3D generative models, recent work has begun exploring training-free 3D editing \cite{liVoxHammerTrainingFreePrecise2026,parelli3DLATTELatentSpace2026,yeNano3DTrainingFreeApproach2026}. However, these methods often apply 2D training-free editing techniques \cite{lugmayrRePaintInpaintingUsing2022,wangTamingRectifiedFlow2025,kulikovFlowEditInversionFreeTextBased2025} to 3D models, still relying on edited 2D views and suffering from the same 3D--2D--3D inconsistency issues. Existing 3D editing approaches also typically require precise 3D masks or well-edited 2D views, which hinder practical usage.

\section{Method}

\begin{figure}[tb!]
    \centering
    \includegraphics[width=\textwidth]{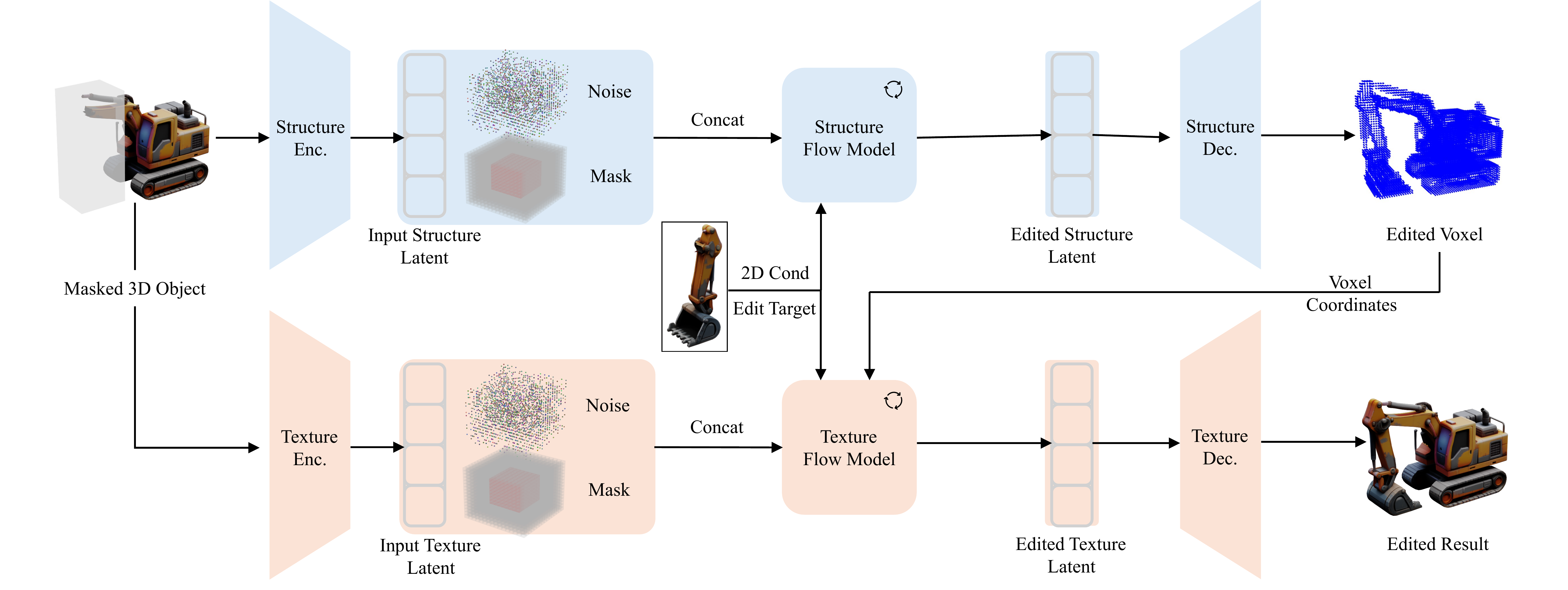}
    \caption{\textbf{An overview of our method.} Given a masked 3D object as input, we first extract its structure and texture latents using the TRELLIS encoder. The input latents are concatenated with a binary mask and random noise along the feature channel dimension, then fed into the flow-matching model, which takes the editing target as a condition. The flow model generates edited latents, which decode into the final result.}
    \label{fig:overview}
\end{figure}

We detail our approach in four stages. First, we describe the selected generative backbone, TRELLIS \cite{xiangStructured3DLatents2025}, in \cref{subsec:preliminary}. Next, we explain how our network architecture supports coarse inputs in \cref{subsec:network architecture}. Third, we introduce the training strategy and data augmentation techniques applied to improve the model's performance in \cref{subsec:training strategy}. Finally, we describe the construction of our large-scale editing dataset in \cref{subsec:dataset curation}. Figure~\ref{fig:overview} shows an overview of our method.

\FloatBarrier

\subsection{Preliminary}
\label{subsec:preliminary}

Our 3D editing model is based on TRELLIS, a 3D generation framework. TRELLIS generates high-quality 3D objects conditioned on either images or text. It uses an encoder-decoder architecture, where the encoder compresses 3D assets into latent representations, and the decoder reconstructs them into various 3D formats, such as 3DGS \cite{kerbl3DGaussianSplatting2023,yuMipSplattingAliasfree3D2024}, NeRF \cite{mildenhallNeRFRepresentingScenes2022}, and Mesh. To enable 3D object generation conditioned on image or text prompts, rectified flow models \cite{lipmanFlowMatchingGenerative2023} are employed. The ground truth for these rectified flow models is the latent representation obtained from encoding the 3D object. During inference, the rectified flow model samples the latent from a noise distribution conditioned on the image or text prompt, and the decoder decodes it into the desired 3D format.

Specifically, TRELLIS handles structure and texture separately. For a given 3D object, TRELLIS first normalizes its vertices into a unit cube within the range of $[-0.5, +0.5]$, then voxelizes it into a $64^3$ voxel grid $p$ ($N \times 3$, $N$ is the number of active voxels) for the structure encoder input. The structure encoder transforms $p$ to a structure latent $L_{s}$ ($16 \times 16 \times 16 \times C_{s}$, $C_{s}$ is the feature channel), which is then decoded back to the voxel form $p$ by the decoder.

For texture, TRELLIS renders 150 2D views and extracts feature maps using the DINOv2 \cite{oquabDINOv2LearningRobust2024,joseDINOv2MeetsText2025,darcetVisionTransformersNeed2024} model. The active voxel coordinates $p$ are projected onto the multi-view feature maps to retrieve features at the corresponding locations. The average of these 2D features $\{z, p\}$ is used as input to the texture encoder, where $z$ is the gathered DINOv2 feature ($N \times C_{d}$, $C_{d}$ is the feature channel of DINOv2 output). The texture encoder encodes $\{z,p\}$ to a texture latent $L_{t}$ ($N \times C_{t}$, $C_{t}$ is the feature channel). The texture decoder then decodes $\{L_{t}, p\}$ into the desired 3D output.

Rectified flow models are introduced to integrate image or text conditions. The flow model follows a forward-backward process similar to that of diffusion models \cite{rombachHighResolutionImageSynthesis2022}. In the forward process, noise $\epsilon$ is added to the data sample $x_{0}$ with $x_{t}=(1-t)x_{0}+t\epsilon$ at each timestep $t$. The backward process is defined as:
\begin{equation}
\frac{dx_{t}}{dt} = v(x_{t}, t),
\end{equation}
which moves noisy samples toward the data distribution. We learn $v$ using a neural network $\theta$ that minimizes the difference between the predicted $v_{\theta}(x_{t}, t)$ and ground-truth vector fields, computed from the forward process:
\begin{equation}
\mathcal{L}(\theta)=\mathbb{E}_{t,x_{0},\epsilon}||v_{\theta}(x_{t},t)-(\epsilon-x_{0})||^{2}_{2}.
\end{equation}
Flow models for structure and texture are trained separately, with the learning targets being the corresponding structure latent $L_{s}$ and texture latent $L_{t}$.

Since TRELLIS's encoder-decoder network is highly effective at compressing and restoring 3D objects with nearly lossless perceptual quality, we directly leverage these models and train the rectified flow models for 3D editing.

\subsection{Network Architecture}
\label{subsec:network architecture}

We encode 3D objects into a latent representation using TRELLIS's encoder, which produces both structure $L_{s}^{in}$ and texture ${L_{t}^{in}}$ latents.

\myheading{Structure.}
For the structure, we concatenate the encoded latent $L_{s}^{in}$ ($16 \times 16 \times 16 \times C_{s}$), the boolean mask $\mathcal{M}_{s}$ ($16 \times 16 \times 16 \times 1$), and noise $\epsilon_{s}$ ($16 \times 16 \times 16 \times C_{s}$) along the feature dimension as the model input $L_{s}^{'}$ ($16 \times 16 \times 16 \times [2C_{s}+1]$). The output is the edited structure latent $L_{s}^{out}$ ($16 \times 16 \times 16 \times C_{s}$). We decode $L_{s}^{out}$ to edited geometry $p^{out}$ ($N^{out} \times 3$) with TRELLIS's decoder.

\myheading{Texture.}
For the texture, we first combine the input 3D object's texture latent ${L_{t}^{in}}$ ($N^{in} \times C_{t}$) and the coordinates of the edited voxels $p^{out}$ ($N^{out} \times 3$) into $\{{L_{t}^{in}}, p^{out}\}$. Note that the input shape to be edited $N^{in}$ and the predicted output shape $N^{out}$ are misaligned. For coordinates that are not covered by the input, we pad the corresponding features with zeros. We then concatenate $\{{L_{t}^{in}}, p^{out}\}$, the boolean mask $\mathcal{M}_{t}$ ($N^{out} \times 1$), and the noise $\epsilon_{t}$ ($N^{out} \times C_{t}$) along the feature dimension, resulting in the edited texture latent $\{{L_{t}^{out}}, p^{out}\}$ ($N^{out} \times [2C_{t}+1]$).

For the condition, we use images with partial elements that indicate the editing target, rather than the complete edited 2D view, for both the structure and texture models. Given an image prompt, we add the encoded latent from the input 3D object and the 3D coarse bounding box, then sample the edited structure and texture latents from noise.

\subsection{Training Strategy}
\label{subsec:training strategy}

\myheading{Region-Aware Loss Reweighting.}
Unlike optimizing a generative model, the editing task inherently involves an imbalance between regions that need editing and those that do not. Areas that do not require editing tend to converge more easily, while the target editing regions incur higher losses that are more difficult to reduce. The loss function for the generation task is simply an MSE loss between the predicted velocity and the ground truth
\begin{equation}
\mathcal{L}_{\mathrm{gen}} \;=\; \frac{1}{n}\sum_{i=1}^{n} \big\|\, v\big(x_t^{(i)},t\big) - v_{\theta}\big(x_t^{(i)},t\big)\,\big\|_2^2.
\end{equation}
Here, $n$ is the number of elements for the flow model's vector field, and $v_{\theta}$ and $v$ are the predicted vector field and ground truth, respectively. For simplicity, we abbreviate $v\big(x_t^{(i)},t\big) - v_{\theta}\big(x_t^{(i)},t\big)$ as $v - v_{\theta}$.

We apply a loss reweighting strategy by normalizing the loss so that both regions (masked and non-masked) receive proportional attention. We scale the loss by the number of elements in each area, ensuring that both masked and non-masked regions contribute equally to the overall loss, even if one region is significantly larger than the other. Suppose $\mathcal{M}$ is the boolean mask, and $m$ and $\bar{m}$ are the number of elements in the masked and nonmasked region, respectively. The loss terms are
\begin{equation}
\mathcal{L}_{m}
= \frac{1}{m} \sum_{i=1}^{m}
\left\|\, v - v_{\theta} \,\right\|_2^2 \, \mathcal{M}_i , \quad
\mathcal{L}_{\bar{m}}
= \frac{1}{\bar{m}} \sum_{i=1}^{\bar{m}}
\left\|\, v - v_{\theta} \,\right\|_2^2 \, (1 - \mathcal{M}_i) .
\end{equation}
Through this loss decomposition, we obtain more balanced loss terms for both the unchanged region and the target editing area.

Moreover, we adopt a hard-example mining strategy to emphasize regions that are more difficult to learn. Specifically, we select the hardest regions corresponding to the top $\tau\%$ of per-index losses:
\begin{equation}
\mathcal{L}_{\text{hard}}
= \frac{1}{|\mathcal{H}|} \sum_{i \in \mathcal{H}}
\left\|\, v - v_{\theta} \,\right\|_2^2 , \quad
\mathcal{H}
= \text{Top-}\tau\%\!\left(
\left\|\, v - v_{\theta} \,\right\|_2^2
\right) ,
\end{equation}
where $\mathcal{H}$ denotes the set of hard examples whose per-index losses rank within the highest $\tau\%$ of all losses.

To balance each loss term's value, the overall loss is formulated as:
\begin{equation}
\mathcal{L}_{\text{edit}}
= \mathcal{L}_{m}
+ \frac{|\mathcal{L}_{m}|}{|\mathcal{L}_{\bar{m}}|} \, \mathcal{L}_{\bar{m}}
+ \frac{|\mathcal{L}_{m}|}{|\mathcal{L}_{\text{hard}}|} \, \mathcal{L}_{\text{hard}} .
\end{equation}

\begin{figure}[tb!]
    \centering
    \includegraphics[width=\textwidth]{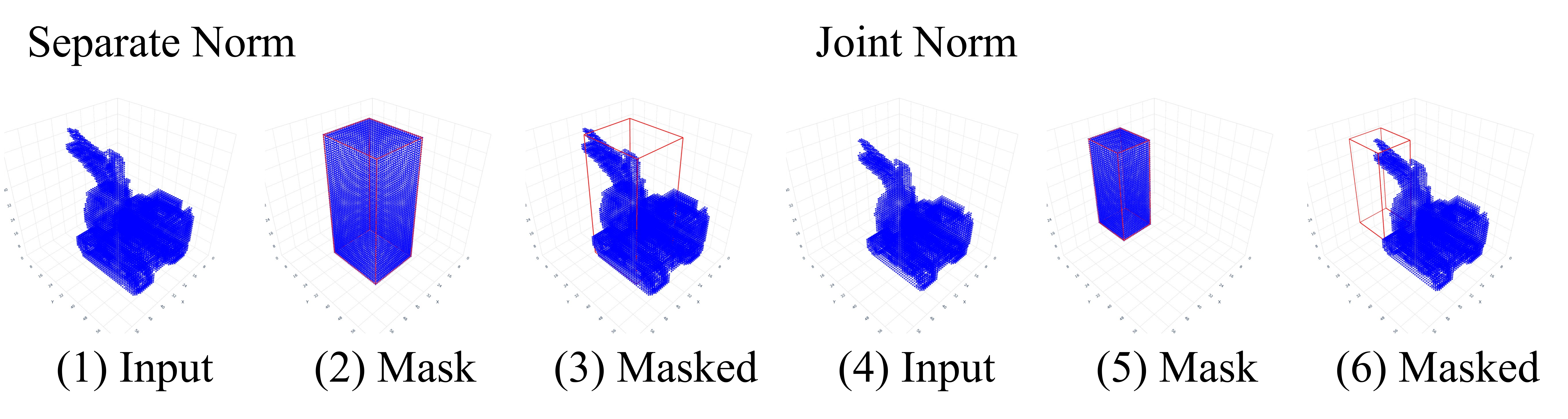}
    \caption{\textbf{Effect of joint normalization.} Separate normalization leads to overlapping (3), whereas our joint normalization preserves the correct spatial relationship (6).}
    \label{fig:normalization}
\end{figure}

\myheading{Joint Normalization for Spatial Alignment.}
A potential issue with the input data is normalization. If the input 3D object and the 3D mask are independently normalized to a unit cube, this can distort their relative positions. We observe that if the 3D object and the mask are not spatially aligned, the model struggles to learn the mapping between the input and the edited output. To address this, we treat the 3D object and the mask as a whole and compute a single normalization factor for both. This normalization factor is then applied uniformly across all relevant content. During inference, we use a similar normalization process to ensure consistency in spatial alignment. Figure~\ref{fig:normalization} shows the necessity of our joint normalization.

\FloatBarrier
\FloatBarrier

\myheading{Training with Coarse 3D Masks.}
In practical scenarios, users prefer to provide a coarse 3D bounding box that is larger than the actual region to be edited rather than defining precise 3D editing masks. To ensure the model can work with the arbitrary coarse 3D mask during inference, we synchronize its use during training. Experiments show that models trained with exact 3D shapes as masks converge well during training but perform poorly when tested with a coarse bounding box. Furthermore, compared to using the minimum bounding box of the accurate 3D mask, applying random disturbances to the box size and position further improves model performance.

\myheading{Filtering Unrealistic Editing Pairs.}
We filter out data pairs in which the target editing region is too small based on the volume occupied by the 3D voxels. This operation aims to ensure that the training data primarily consists of cases likely to occur in real editing scenarios. Target regions with tiny volumes often correspond to editing pairs that lack practical significance. The effectiveness of this data filtering strategy is further validated in our experimental results. Please refer to the supplementary materials for more details.

\begin{figure}[H]
    \centering
    \includegraphics[width=\textwidth]{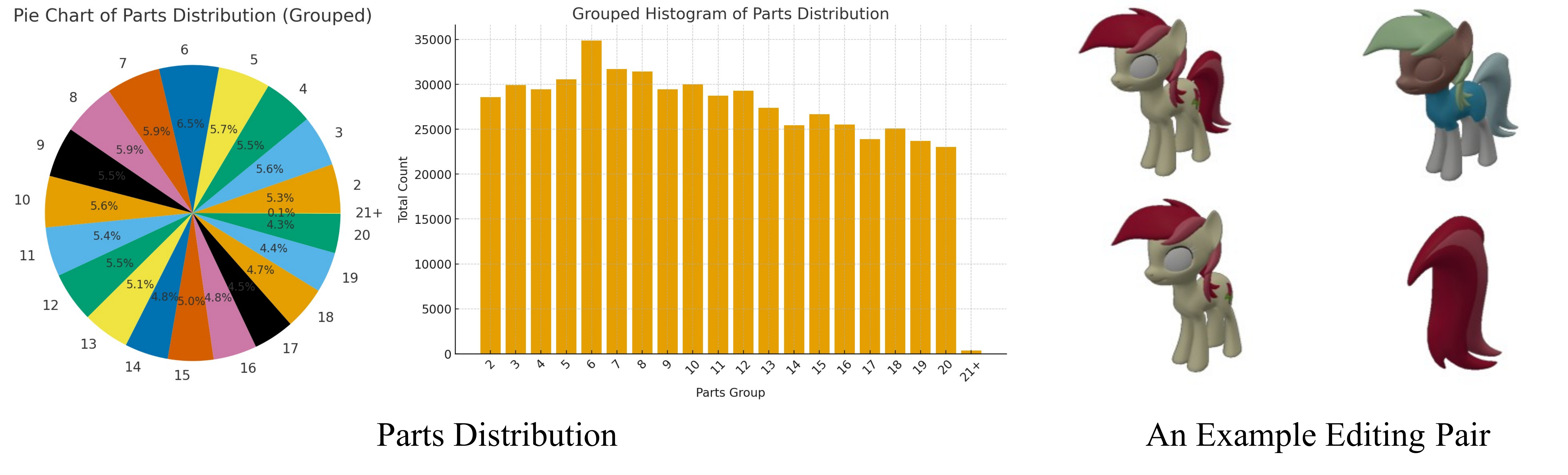}
    \caption{\textbf{Overview of our 3D editing dataset.} Left: \textit{Parts Distribution} across the dataset. Right: \textit{An Example Editing Pair} illustrating the ground truth (top left), segmentation (top right), source object (bottom left), and editing target (bottom right).}
    \label{fig:dataset}
\end{figure}
\FloatBarrier

\subsection{Dataset Curation}
\label{subsec:dataset curation}

To train our model end-to-end, we create a large-scale dataset of 3D editing pairs. The basic idea is that, for a given 3D object, we can treat its partially deleted version as the object to be edited, with the original 3D object serving as the ground truth. These two objects form an editing pair. However, directly deleting parts of the 3D object can result in missing information at cut or fracture regions, hindering model performance. To address this, we collect 3D objects with complete part information whose distinct parts still preserve a whole structure and are suitable for constructing editing pairs.

We organize our dataset using manually validated 3D segmentation data from the Partverse dataset \cite{dongOneMoreContextual2025} and 3D assets with part information from the Objaverse dataset \cite{deitkeObjaverseUniverseAnnotated2023,deitkeObjaverseXLUniverse10M2023}. We clean the data by removing low-quality instances (e.g., geometric distortions or missing textures). In the end, we construct a dataset comprising approximately 85k meshes and 500k editing pairs. We compute bounding boxes and generate image prompts from the deleted 3D parts. Figure~\ref{fig:dataset} shows an overview of our dataset.

\section{Experiments}

\subsection{Implementation}

\myheading{Train.} For each editing pair in our dataset, we use the TRELLIS encoder to extract the structure and texture latents of the input 3D object, which serve as inputs to the flow-matching model. We compute the minimum 3D bounding box of the target 3D part and apply our box augmentation strategy to formulate the 3D mask for training. Subsequently, we render 24 views of the 3D part from cameras uniformly sampled across a bounding sphere. At each training step, one of these views is randomly selected to act as the condition. The flow-matching models for structure and texture are trained independently, each for 10k steps with a batch size of 1 on 96 V100 (32GB) GPUs. The complete two-stage training process requires approximately 10k GPU hours.

\myheading{Evaluation.} We use Chamfer Distance (CD) \cite{fanPointSetGeneration2017} to measure geometry quality and render 32 views to assess texture quality using PSNR, SSIM \cite{wangImageQualityAssessment2004}, LPIPS \cite{zhangUnreasonableEffectivenessDeep2018}, DINO feature similarity \cite{oquabDINOv2LearningRobust2024}, and FID \cite{heuselGANsTrainedTwo2017}. We build a test set of about 200 meshes and 1500 3D editing pairs from the PartObjaverse-Tiny \cite{yangSAMPart3DSegmentAny2024} dataset, following our data generation pipeline, serving as the benchmark for adding. Additionally, we use the VoxHammer \cite{liVoxHammerTrainingFreePrecise2026} dataset, which contains 100 meshes and 300 editing pairs, as an evaluation set for replacement. Since this dataset lacks ground-truth edits, we compute quantitative metrics only on non-edited regions. To simulate real-world scenarios in which users may not provide accurate bounding boxes, we use bounding boxes of varying sizes and positions during evaluation, rather than those used in training. Note that during training, each editing pair is associated with 24 condition views. During evaluation, to assess the robustness of our model against image prompts from varying viewing angles, we report the average performance across all 24 views.

\subsection{Main Results}

We compare our method to Instant3dit \cite{bardaInstant3ditMultiviewInpainting2025}, TRELLIS \cite{xiangStructured3DLatents2025} (two re-implemented local editing variants), and VoxHammer \cite{liVoxHammerTrainingFreePrecise2026}.
Instant3dit renders four canonical views, applies text-driven edits on those renderings, and reconstructs a 3D result from the four edited views. For TRELLIS, we re-implement two local editing strategies: \textbf{(1)} Repaint \cite{lugmayrRePaintInpaintingUsing2022} — which mixes the model-predicted target region with the preserved region (the input injected with noise) at each inference timestep, and \textbf{(2)} FlowEdit \cite{kulikovFlowEditInversionFreeTextBased2025} — which directly drives the input toward the target prompt by computing a new vector field guided by both the source and target prompts. VoxHammer adapts RF-Edit \cite{wangTamingRectifiedFlow2025} to TRELLIS, achieving more accurate inversion and applying Repaint not only to the latent representation but also to intermediate-layer features.

\myheading{Quantitative.} Our method outperforms existing approaches on both replacement and addition tasks (\cref{tab:main results edits}), in terms of both edit-region fidelity and preservation of unedited regions (\cref{tab:main results regions}). \textbf{(1)} Instant3dit: Its 2D editing capability is limited, and its 2D-to-3D process relies solely on reconstruction from four sparse canonical views rather than dense views, and lacks mechanisms to resolve 3D inconsistencies. These factors contribute to a significant performance degradation, resulting in the worst quantitative results among all evaluated baselines. \textbf{(2)} Repaint \& FlowEdit: Repaint significantly outperforms FlowEdit. This superiority stems from Repaint's mask-based fusion, which leverages the unedited region to constrain the generated content within the target editing area. In contrast, FlowEdit predicts vector fields corresponding to the source and target prompts to compute the editing direction. This guidance mechanism is relatively weak. \textbf{(3)} VoxHammer: By utilizing intermediate network features, it should achieve better performance than the direct application of Repaint. However, based on the authors' implementation, we observed slightly inferior results. This discrepancy may arise from differences in our specific evaluation configurations.

\begin{table}[H]
\centering
\caption{\textbf{Quantitative comparison of different edits.} \textit{Repaint} and \textit{FlowEdit} denote two 2D editing approaches applied to TRELLIS.
For \textit{Replace}, since ground truth is unavailable for edited regions, the results shown correspond only to the preserved regions. For \textit{Add}, we report metrics across the entire edited results.}
\label{tab:main results edits}
\resizebox{\textwidth}{!}{
\begin{tabular}{c|cccccc|cccccc}
\toprule
\multirow{2}{*}{Method} & \multicolumn{6}{c|}{Replace} & \multicolumn{6}{c}{Add} \\
& CD$\downarrow$ & PSNR$\uparrow$ & SSIM$\uparrow$ & LPIPS$\downarrow$ & DINO$\uparrow$ & FID$\downarrow$ & CD$\downarrow$ & PSNR$\uparrow$ & SSIM$\uparrow$ & LPIPS$\downarrow$ & DINO$\uparrow$ & FID$\downarrow$ \\
\midrule
Instant3dit  & 0.297 & 13.06 & 0.868 & 0.255 & 0.732 & 73.86 & 0.110 & 8.075 & 0.664 & 0.501 & 0.681 & 81.78 \\
Repaint      & 0.007 & 35.96 & 0.992 & 0.010 & 0.980 & 29.32 & 0.008 & 27.44 & 0.960 & 0.038 & 0.978 & 3.408 \\
FlowEdit     & 0.071 & 17.93 & 0.917 & 0.109 & 0.902 & 40.76 & 0.017 & 19.46 & 0.886 & 0.102 & 0.945 & 6.129 \\
VoxHammer    & 0.022 & 24.51 & 0.961 & 0.043 & 0.955 & 36.28 & 0.023 & 21.16 & 0.923 & 0.095 & 0.919 & 8.281 \\
\midrule
Ours         & \textbf{0.005} & \textbf{36.32} & \textbf{0.995} & \textbf{0.008} & \textbf{0.981} & \textbf{28.34} & \textbf{0.005} & \textbf{28.67} & \textbf{0.962} & \textbf{0.029} & \textbf{0.984} & \textbf{2.960} \\
\bottomrule
\end{tabular}
}
\end{table}

\begin{table}[H]
\centering
\caption{\textbf{Quantitative comparison for different regions}. Since some texture metrics (e.g., LPIPS, DINO feature similarity) do not support irregularly shaped images, the \textit{Target Editing Region} is computed with the minimal 2D bounding box. The \textit{Unedited Region} is evaluated by treating the edited region as consistent with the ground truth. The above estimation yields a higher metric for actual performance but does not compromise consistent comparison. Because the \textit{Target Editing Region}'s sizes vary, FID is not reported because it does not support inputs of different sizes.}
\label{tab:main results regions}
\resizebox{\textwidth}{!}{
\begin{tabular}{c|ccccc|ccccc}
\toprule
\multirow{2}{*}{Method} & \multicolumn{5}{c|}{Target Editing Region} & \multicolumn{5}{c}{Unedited Region} \\
& CD$\downarrow$ & PSNR$\uparrow$ & SSIM$\uparrow$ & LPIPS$\downarrow$ & DINO$\uparrow$ & CD$\downarrow$ & PSNR$\uparrow$ & SSIM$\uparrow$ & LPIPS$\downarrow$ & DINO$\uparrow$ \\
\midrule
Instant3dit & 0.052 & 3.698 & 0.119 & 0.194 & 0.406 & 0.105 & 8.529 & 0.746 & 0.425 & 0.660 \\
Repaint     & 0.019 & 19.00 & 0.726 & 0.032 & 0.862 & 0.006 & 31.16 & 0.945 & 0.012 & 0.983 \\
FlowEdit    & 0.020 & 14.31 & 0.512 & 0.055 & 0.783 & 0.022 & 23.01 & 0.931 & 0.057 & 0.952 \\
VoxHammer   & 0.039 & 14.26 & 0.499 & 0.087 & 0.696 & 0.019 & 25.86 & 0.964 & 0.044 & 0.948 \\
\midrule
Ours        & \textbf{0.009} & \textbf{21.01} & \textbf{0.811} & \textbf{0.030} & \textbf{0.899} & \textbf{0.002} & \textbf{35.45} & \textbf{0.987} & \textbf{0.008} & \textbf{0.990} \\
\bottomrule
\end{tabular}
}
\end{table}

\begin{figure}[p]
\centering
    \includegraphics[width=\textwidth]{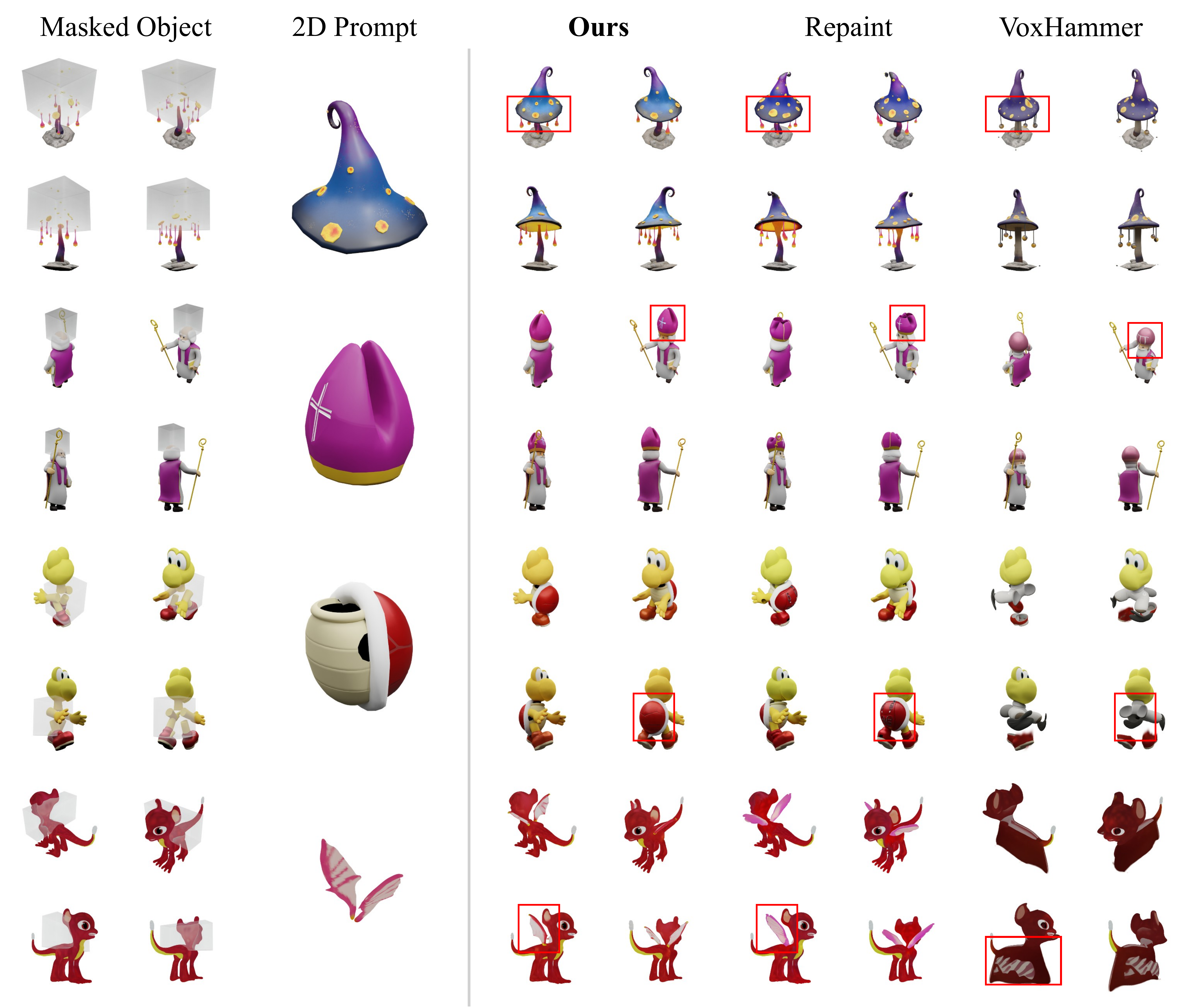}
    \caption{\textbf{Qualitative comparison.} Our method yields prompt-aligned edits while preserving unedited regions. \textit{Repaint} aligns structures in rows (1, 4) but misaligns textures in (2, 3). \textit{VoxHammer} shows partial consistency in (1, 2) but fails in (3, 4).}
    \label{fig:comparison}
\medskip
\centering
    \includegraphics[width=\textwidth]{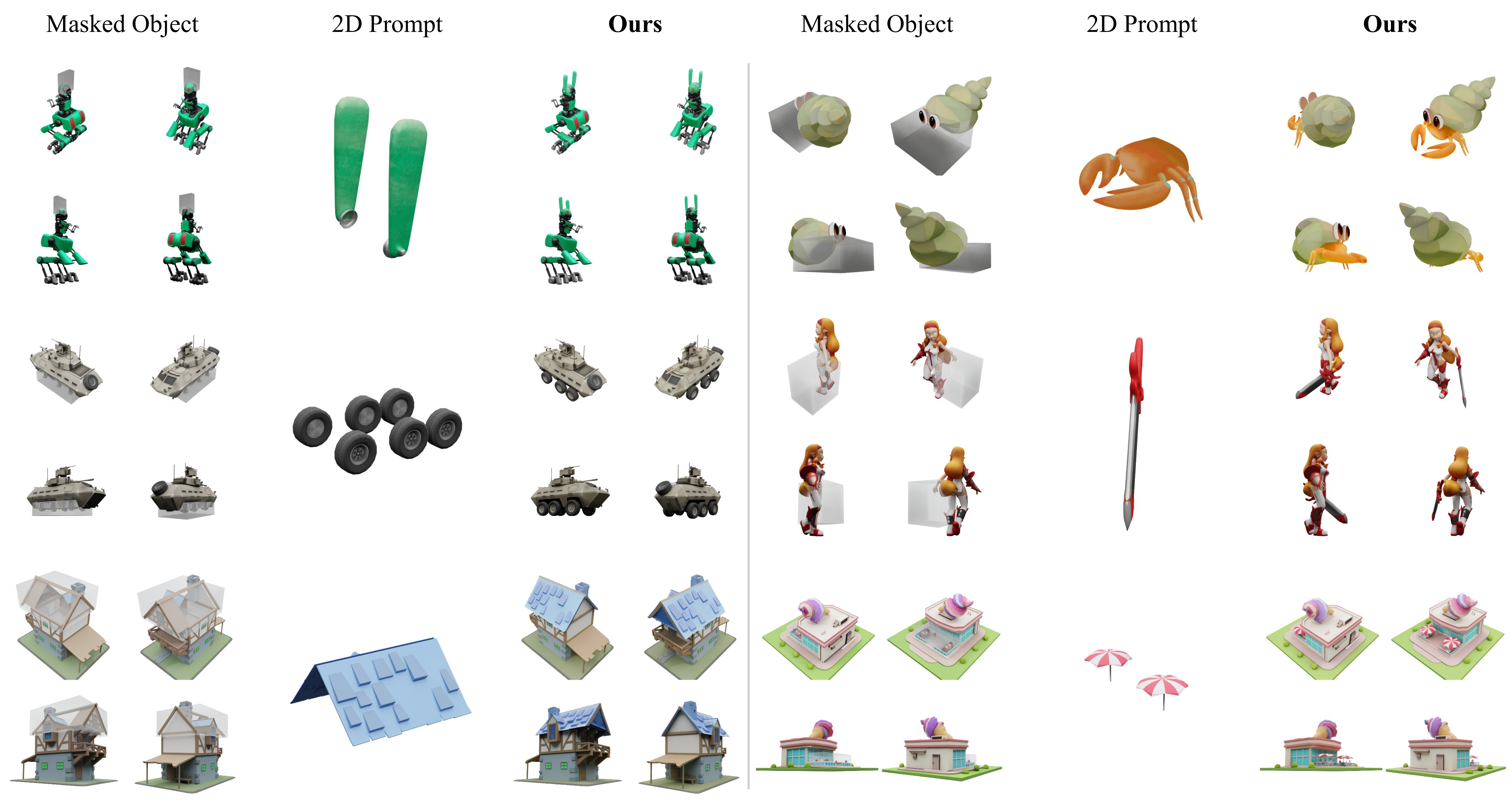}
    \caption{\textbf{Qualitative results.} See additional examples in the supplementary material.}
    \label{fig:show}
\end{figure}

\myheading{Qualitative.} We visualize the editing results produced by our method and existing approaches in \cref{fig:comparison}. As shown, our method delivers clearly superior visual quality. Additional results are presented in \cref{fig:show}, demonstrating the strong generalization ability and robustness of our approach. Across diverse test cases, our method consistently produces realistic and high-quality 3D editing results.

\FloatBarrier

\subsection{Ablation Study}

To validate our method, we perform a detailed ablation study on the PartObjaverse Tiny test set. Variant without \textit{Joint Normalization} fails to converge and is thus omitted. Beyond the components discussed in \cref{subsec:training strategy}, we also examine the impact of data \textit{Segmentation Quality} and training conditions' \textit{View Variance}.  \cref{tab:ablation study} shows that our training strategy consistently improves performance, with Exp.\#5 achieving the best results.
Notably, segmentation quality has a limited impact at the same data scale (Exp.\#5 vs Exp.\#6), indicating that 3D assets with partial information are usable, enabling us to construct a large-scale dataset despite the scarcity of 3D segmentation datasets.

\myheading{Region-Aware Loss Reweighting.} As shown in \cref{tab:ablation study}, our improved loss function (Exp.\#5) outperforms the vanilla generation MSE loss (Exp.\#4).

\myheading{Training with Coarse 3D Masks.} We observe that as the training mask progresses from the exact mask shape (Exp.\#1) to the minimal enclosing bounding box (Exp.\#2), and further to the dilated bounding box (Exp.\#3), model performance steadily improves. Notably, although training with the exact mask shape achieves good results on the training set, it generalizes poorly to the test set and produces meaningless outputs.

\myheading{Filtering Unrealistic Editing Pairs.} Filtering out tiny abnormal instances (Exp.\#4) improves performance compared to the unfiltered dataset (Exp.\#3).

\myheading{Segmentation Quality.} Regarding data sources, datasets with segmentation annotations yield higher-quality segmentation (Exp.\#5) than 3D assets with

\FloatBarrier

\begin{table}[H]
\centering
\caption{\textbf{Ablation study of our method.} The \textit{Mask} column indicates the scale of the 3D mask used during training: \textit{Exact} uses the precise shape, \textit{BBox} uses the minimal enclosing bounding box, and \textit{BBox$^{+}$} applies additional perturbations to simulate real-world scenarios. \ding{51}/\ding{55} denotes whether a data filter is applied. \textit{MSE} and \textit{Ours} refer to the original generation loss and our optimized loss, respectively. \textit{Seg} indicates the quality of data segmentation, with \textit{High} representing human-annotated 3D segmentation datasets and \textit{Low} representing 3D assets with built-in part information. \textit{Cond} refers to the strategy for selecting 2D conditions during training described by \textit{View Variance}. The CD metric is scaled by $10^{2}$ for better visualization.}
\label{tab:ablation study}
\resizebox{\textwidth}{!}{
\begin{tabular}{c|ccccc|cccccc}
\toprule
Exps & Mask & Filter & Loss & Seg & Cond & CD$\times10^{2}$$\downarrow$ & PSNR$\uparrow$ & SSIM$\uparrow$ & LPIPS$\downarrow$ & DINO$\uparrow$ & FID$\downarrow$\\
\midrule
\#1 & Exact         & \ding{55} & MSE   & High & random     & 2.502 & 21.03 & 0.918 & 0.113 & 0.902 & 9.329 \\
\#2 & BBox          & \ding{55} & MSE   & High & random     & 0.701 & 24.65 & 0.946 & 0.057 & 0.960 & 3.709 \\
\#3 & BBox$^{+}$    & \ding{55} & MSE   & High & random     & 0.692 & 24.75 & 0.946 & 0.056 & 0.962 & 3.573 \\
\#4 & BBox$^{+}$    & \ding{51} & MSE   & High & random     & 0.658 & 24.82 & 0.947 & 0.055 & 0.962 & 3.542 \\
\#5 & BBox$^{+}$    & \ding{51} & Ours  & High & random     & \textbf{0.635} & \textbf{25.01} & \textbf{0.947} & \textbf{0.054} & \textbf{0.963} & 3.496 \\
\#6 & BBox$^{+}$    & \ding{51} & Ours  & Low  & random     & 0.636 & 24.90 & 0.947 & 0.054 & 0.963 & \textbf{3.474} \\
\#7 & BBox          & \ding{55} & MSE   & High & traverse   & 0.850 & 24.41 & 0.946 & 0.059 & 0.953 & 4.129 \\
\bottomrule
\end{tabular}
}
\end{table}
\noindent part information (Exp.\#6). Experiments show that models trained on these two sources achieve comparable performance.

\FloatBarrier

\myheading{View Variance.} In real-world scenarios, the orientation of the 3D object and the 2D prompt are often misaligned. To improve generalization to the 2D prompt's view angle, we render 24 views from the partial 3D object and randomly select one per training iteration. Comparing this random selection strategy (Exp.\#2) with traversing all 24 renderings (Exp.\#7) shows that the random approach performs better, a counterintuitive result.

\begin{figure}[!t]
\centering
    \includegraphics[width=\textwidth]{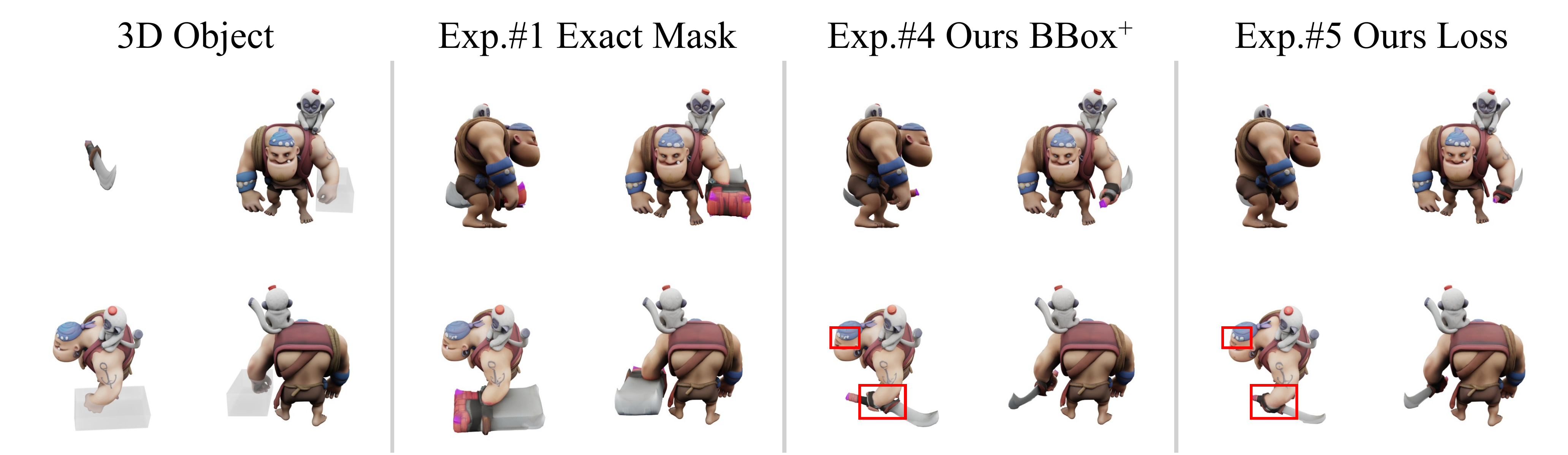}
    \caption{\textbf{Qualitative ablation results.} \textit{3D Object} displays the masked object alongside the target prompt (a sword). \textit{Exact Mask} (Exp.\#1) is trained with precise 3D masks. \textit{Ours BBox$^+$} (Exp.\#4) utilizes our mask augmentation strategy. \textit{Ours Loss} (Exp.\#5) further incorporates our region-aware loss.}
    \label{fig:ablation_study}
\medskip
\centering
    \includegraphics[width=\textwidth]{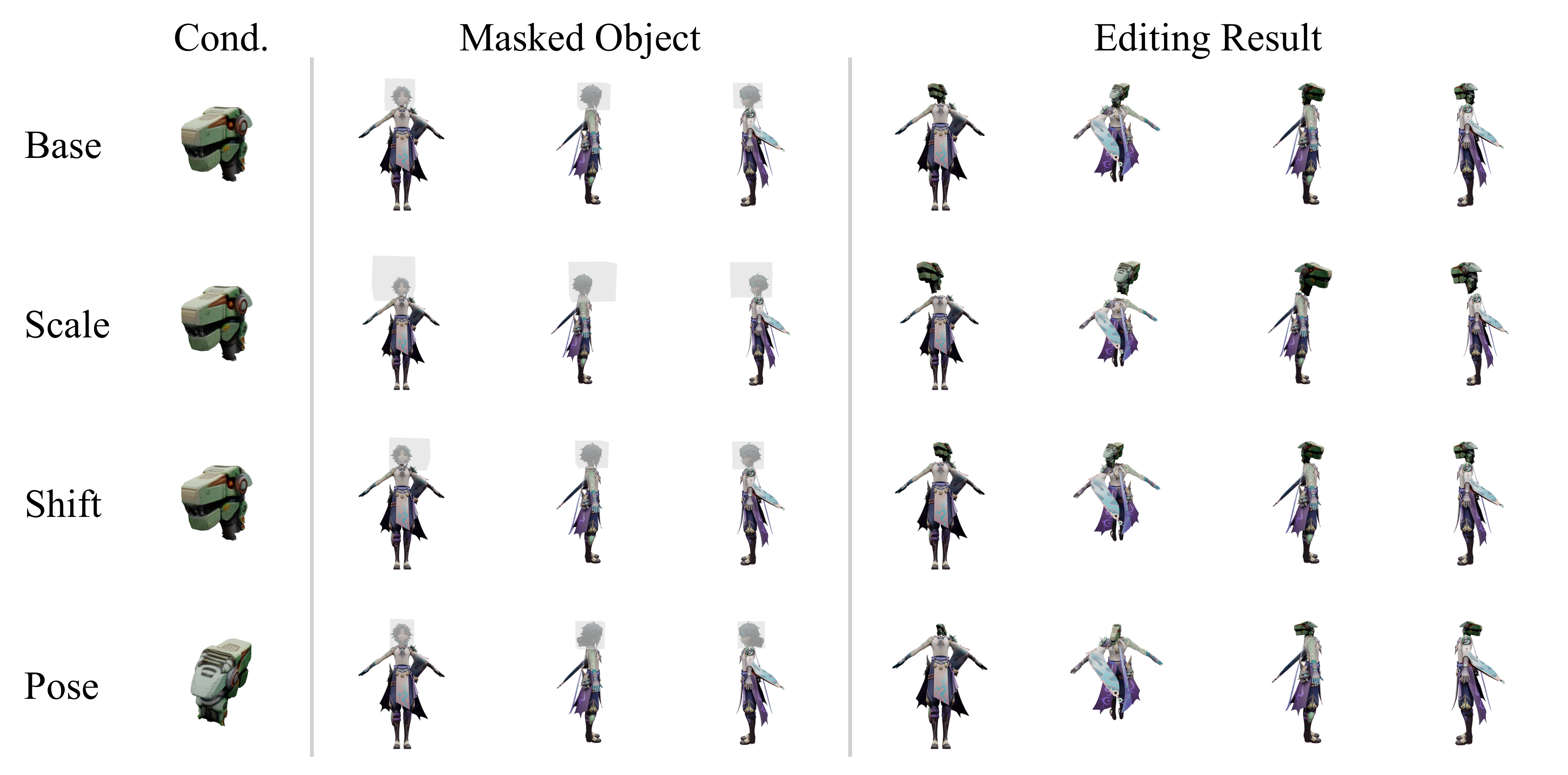}
    \caption{\textbf{Robustness to input variations.} \textit{Base} shows editing results using an ideal bounding box and a 2D prompt. \textit{Scale} enlarges the baseline box by 1.5$\times$. \textit{Shift} translates the scaled box to the right. Finally, \textit{Pose} demonstrates results when the image prompt (Cond.) features a different viewing angle than the \textit{Base}.}
    \label{fig:input_variance}
\end{figure}

\myheading{Quality.} Figure~\ref{fig:ablation_study} shows that exact 3D masks yield blurred geometry in the target area, whereas our mask augmentation strategy significantly improves visual quality. Furthermore, our region-aware loss refines fine-grained details. It preserves unedited attributes (e.g., hat dots, arm tattoos). It enhances consistency between the image prompt and the edited geometry, particularly at transitional boundaries (e.g., the sword's hilt-to-blade junction).

\myheading{Robustness to Input Variance.} Figure~\ref{fig:input_variance} demonstrates the robustness of our method against input perturbations. Our approach consistently produces plausible 3D editing results, despite variations in the positions and scales of the coarse 3D bounding boxes and in the viewing angles of the image prompt.

\subsection{Inference Efficiency}

\cref{tab:efficiency} presents a comparison of inference efficiency between our method and the baselines. \textbf{(1)} Instant3dit: It is slow due to inefficient 3D reconstruction and impractical because it demands exact 3D masks. \textbf{(2)} Repaint \& FlowEdit: The modifications applied at each inference timestep do not introduce computational overhead, thereby maintaining the efficiency of vanilla TRELLIS. However,

\FloatBarrier

\noindent utilizing resampling steps repetition introduces a trade-off between generation quality and inference speed. Regarding inputs, both variants require a pre-edited 2D view as guidance. Specifically, while Repaint does not strictly mandate an exact mask, providing one yields higher-quality results by ensuring the faithful preservation of larger unedited regions. Conversely, FlowEdit operates entirely without a mask, but this lack of spatial constraint renders its inference trajectory highly uncontrollable. \textbf{(3)} VoxHammer: It exhibits a significantly longer inference time due to the computational burden of integrating inner-layer features. Furthermore, it shares Repaint's stringent input requirements, putting it at a distinct disadvantage compared to our approach. In contrast, \textbf{(4)} our proposed method operates as a streamlined, end-to-end framework, preserving inference efficiency comparable to the vanilla TRELLIS while reducing the input burden.

\FloatBarrier
\begin{table}[H]
\centering
\caption{\textbf{Efficiency Comparison.} \textit{Exact} and \textit{Bbox$^+$} indicate whether the method requires an accurate 3D mask or only a coarse 3D bounding box, respectively. A \ding{51} denotes that the resource is required, while a \ding{55} indicates that it is not.}
\label{tab:efficiency}
\resizebox{\textwidth}{!}{
\begin{tabular}{l|c|cccc|c}
\toprule
Method      & TRELLIS & Instant3dit & Repaint   & FlowEdit  & VoxHammer & Ours \\
\midrule
3D Mask     & --      & Exact       & BBox$^+$  & \ding{55} & Exact     & BBox$^+$ \\
2D Mask     & --      & \ding{51}   & \ding{55} & \ding{55} & \ding{51} & \ding{55} \\
Edited 2D   & --      & \ding{51}   & \ding{51} & \ding{51} & \ding{51} & \ding{55} \\
Runtime(s)  & 20      & 30          & 20        & 20        & 120       & 20 \\
\bottomrule
\end{tabular}
}
\end{table}

\FloatBarrier
\section{Conclusion}
\label{sec:conclusion}

In this work, we propose a novel end-to-end pipeline for local 3D object editing, which takes as input a 3D object, a coarse bounding box, and a reference 2D image. Our approach achieves high-fidelity results by introducing region-aware adaptive loss and data augmentation techniques. Extensive experiments demonstrate that our approach achieves state-of-the-art performance, outperforming existing methods in both editing quality and efficiency. Furthermore, we construct a large-scale dataset of 3D editing pairs, facilitating supervised training and future research in 3D editing.

\section*{Acknowledgements}
This research is supported by the MoE AcRF Tier 2 grant (MOE-T2EP20223-0001) and the MoE AcRF Tier 1 grant (RG14/22). This research is also supported by a Tencent research grant (04IDS001541N022).

\bibliographystyle{splncs04}
\bibliography{eccv2026}

\end{document}